\documentclass[applsci,review,accept,pdftex,moreauthors]{Definitions/mdpi} 

\usepackage{array}
\usepackage{multirow}
\usepackage{booktabs}
\usepackage{adjustbox}
\usepackage{tabularx}
\usepackage{graphicx} 
\usepackage{makecell}

\firstpage{1} 
\pubvolume{1}
\issuenum{1}
\articlenumber{0}
\pubyear{2025}
\copyrightyear{2025}
\externaleditor{Firstname Lastname} 
\datereceived{22 December 2024} 
\daterevised{19 March 2025} 
\dateaccepted{21 March 2025} 
\datepublished{ } 
\hreflink{https://doi.org/} 

\Title{Neural 
 Architecture Search for Generative Adversarial Networks: A Comprehensive Review and Critical Analysis}

\TitleCitation{Neural Architecture Search for Generative Adversarial Networks: A Comprehensive Review and Critical Analysis}

\Author{Abrar Alotaibi 
 $^{1,2,}$*$^{}$
\orcidA{}, Moataz Ahmed $^{1,3}$\orcidB{}}

\AuthorNames{Abrar Alotaibi and Moataz Ahmed}

 \AuthorCitation{Alotaibi, A.; Ahmed, M.}

\address{%
$^{1}$ \quad Department of Information and Computer Science, King Fahd University of Petroleum and Minerals, Dhahran,31261, 
 Saudi Arabia  
\\
$^{2}$ \quad Computer Science Department, College of Computer Science and Information Technology, Imam Abdulrahman bin Faisal University, Dammam, 31441, Saudi Arabia\\
$^{3}$ \quad SDAIA-KFUPM Joint Research Center for Artificial Intelligence, King Fahd University of Petroleum and Minerals, Dhahran, 31261, Saudi Arabia; moataz@kfupm.edu.sa}

\corres{Correspondence: amotaibi@iau.edu.sa}

\abstract{Neural Architecture Search (NAS) has emerged as a pivotal technique in optimizing the design of Generative Adversarial Networks (GANs), automating the search for effective architectures while addressing the challenges inherent in manual design. This paper provides a comprehensive review of NAS methods applied to GANs, categorizing and comparing various approaches based on criteria such as search strategies, evaluation metrics, and performance outcomes. The review highlights the benefits of NAS in improving GAN performance, stability, and efficiency, while also identifying limitations and areas for future research. Key findings include the superiority of evolutionary algorithms and gradient-based methods in certain contexts, the importance of robust evaluation metrics beyond traditional scores like Inception Score (IS) and Fréchet Inception Distance (FID), and the need for diverse datasets in assessing GAN performance. By presenting a structured comparison of existing NAS-GAN techniques, this paper aims to guide researchers in developing more effective NAS methods and advancing the field of GANs.}

\begin{document}

\section{Introduction}
\unskip

\subsection{Background}
Data insufficiency poses a recurring challenge for many classification tasks, hampering both model development and validation. With~limited data, models can struggle to learn robust feature representations, leading to poor generalization on unseen data~\cite{shorten2019survey,souza2021fine}. This problem is compounded when working with complex models like deep neural networks that have high capacity and are prone to overfitting~\cite{elaziz2021advanced}. To~address data scarcity, recent advancements have turned to data augmentation (DA), which involves creating new training samples from existing ones. DA is a powerful technique for addressing both limited datasets and class imbalance. By~generating additional examples from rare classes, oversampling with DA can rebalance class distributions. Moreover, augmentation helps reduce overfitting when training deep neural networks (DNNs) by expanding the training set with transformed versions of existing samples. This exposes models to greater diversity without needing to collect new~data.

DA techniques have been significantly empowered using generative adversarial networks (GANs) \cite{shorten2019survey}. GANs can learn complex data distributions and generate realistic synthetic samples for augmentation. They have become a pivotal DA tool, especially in areas like medical imaging, anomaly detection, and~text-to-image generation~\cite{apostolopoulos2022applications,xia2022gan,kocasari2022stylemc}. GAN-based DA is quickly becoming indispensable for robust deep learning with scarce data across modalities like images, video, audio, and~text~\cite{shorten2019survey,kocasari2022stylemc,mosolova2018text}.

\subsection{Motivation}
Developing GANs poses some considerable challenges. Like many deep learning techniques, GANs require meticulous hyperparameter tuning and network architecture selection to optimize results~\cite{talbi2020optimization}. The~model hyperparameters and structure heavily influence the final generated samples. Manually tuning these complex generative models with dozens of hyperparameters is often tedious and suboptimal, and~requires expert knowledge. Additionally, the~simultaneous training of the generator and discriminator networks makes GANs notoriously unstable and difficult to converge~\cite{souza2021fine}. There can be misleading situations where the discriminator is unable to properly judge the quality of the generated images, rather than the generator producing high-quality artificial samples. This imbalance prevents effective adversarial learning. Moreover, problems like mode collapse may arise where the generator lacks diversity and produces limited varieties of samples. Careful regularization and architectural choices are needed to promote generator diversity~\cite{thanh2020catastrophic}. Researchers have put considerable effort into manually improving GAN architectures, but~this process demands significant~expertise.

Recently, Neural Architecture Search (NAS) has emerged as an effective tool for automatically discovering superior models across various tasks, including GANs~\cite{elsken2019neural,he2021automl}. Early attempts at applying NAS to GANs focused solely on optimizing the generator while keeping the discriminator fixed, to~simplify the search process~\cite{gong2019autogan}. However, this approach may result in suboptimal GANs. More recent studies have attempted to search for both generator and discriminator architectures simultaneously, but~they face challenges due to the inherent instability of GAN training~\cite{costa2019coegan,tian2021alphagan}.

\subsection{Objectives}
In our research, we have developed a framework to categorize and compare different techniques based on a set of key criteria identified through an extensive review of existing methods. Using this framework, we conducted a critical analysis and comparison of the various NAS-GAN techniques currently available in the literature. This assessment not only allowed us to evaluate existing approaches but also highlighted areas for potential future~research.

The remainder of this paper is structured as follows: Section~\ref{sec2} reviews related work. In~Section~\ref{sec3}, we describe the research methodology and outline the research questions. Section~\ref{sec4} provides the analysis, results, and~responses to the research questions. Section~\ref{sec5} discusses the study's implications and proposes directions for future research. Section~\ref{sec6} addresses the potential threats to the validity of this study. Finally, Section~\ref{sec7} concludes the~paper.

\section{Related~Studies}\label{sec2}
\unskip

\subsection{Overview of NAS-GAN~Reviews}
Despite the growing interest in applying Neural Architecture Search (NAS) to Generative Adversarial Networks (GANs), there have been limited comprehensive reviews on this specific topic. To~date, only two papers have been published that specifically review NAS in GANs, and~these are essentially one work and its~continuation.

Ganepola and Wirasingha (2021) \cite{ganepola2021automating} conducted a comprehensive review of NAS techniques applied to GANs. The~authors analyzed various approaches based on key components of NAS: search space, search strategy, and~performance estimation strategy. They identified cell-based and chain/entire structure as the primary search space types, with~reinforcement learning, gradient-based, and~evolutionary algorithms as the main search strategies. The~review focused on image generation and GAN model compression tasks, comparing different methods using metrics such as Inception Score (IS) and Fréchet Inception Distance (FID). The~authors also discussed limitations and future directions for NAS in GANs, including potential applications in semantic image segmentation and high-resolution image synthesis. However, this survey was published in 2021 and may not include the most recent advancements in the~field.

Buthgamumudalige and Wirasingha (2021) \cite{buthgamumudalige2021neural}, published in the same year, provides a comprehensive review of NAS techniques applied to GANs for image generation tasks. The~authors analyze various approaches using multiple criteria, including search space design, search strategy implementation, and~performance evaluation methods. They examine systems that weren't discussed in the previous paper, comparing them across several key dimensions: image generation quality, computational efficiency, transferability to different datasets, and~support for supervised vs. unsupervised learning. Performance is evaluated using metrics like IS and FID on datasets such as CIFAR-10 and STL-10. The~review also considers practical aspects such as GPU costs and training times. The~paper highlights the progress made in automating GAN architecture design, noting improvements in image quality and search efficiency. However, it also identifies limitations in existing work, such as the narrow focus on specific datasets and image generation types. The~review concludes by suggesting potential areas for future research, emphasizing the need for more comprehensive evaluation criteria, including performance on diverse datasets and exploration of conditional and semi-supervised image generation~tasks.

In contrast, the~survey by Wang~et~al. (2024) \cite{article} presents an extensive analysis of evolutionary computation (EC) applied to GANs. It delves into technical aspects such as the design of mutation operators, the~formulation of fitness functions based on metrics like inverted generational distance (IGD), and~strategies to alleviate mode collapse through Pareto front approximations. The~survey also examines scalability—demonstrating how evolved architectures can transfer from low-dimensional (n 
 = 10) to high-dimensional (\mbox{n = 784}) settings—and discusses integrating EC with gradient-based methods to enhance convergence and robustness. Although~its scope is broad, covering NAS, parameter tuning, loss function adaptation, and~synchronization strategies, the~survey is best categorized within the wider NAS reviews rather than a strictly NAS-GAN-focused~review.

\subsection{Other NAS Reviews with GAN~Sections}
Other reviews that were focused on NAS methods had a dedicated section for GANs. Kang~et~al.'s (2023) \cite{kang2023neural} review focused on the application of NAS in computer vision tasks, dividing them into detection, segmentation, and~generation. It presented prominent works that applied NAS to GANs for the generation task. On~the other hand, White~et~al. (2023)~\cite{white2023neural} extensively reviewed NAS, garnering insight from 1000 papers. The~GAN section of the review focused on identifying the best techniques used amongst the prominent works. As~summarized in Table~\ref{tab:review-comparison}, these works either lacked comprehensive coverage of NAS-GAN methodologies or focused narrowly on specific tasks. In~contrast, our review synthesizes advancements from 2021–2025, addressing gaps such as dataset diversity and reproducibility while maintaining a GAN-specific~focus.

\begin{table}[H] 
\small 
\caption{Comparison 
 of NAS-GAN Review~Papers.}
\label{tab:review-comparison} 

\begin{adjustwidth}{-\extralength}{0cm}
 \begin{tabularx}{\fulllength}{m{3.1cm}<{\raggedright}m{0.6cm}<{\raggedright}m{4.1cm}<{\raggedright}m{4.1cm}<{\raggedright}X}
  \toprule 
  \textbf{Review Paper} & \textbf{Year} & \textbf{Scope} & \textbf{Limitations} & \textbf{This Review’s Improvements} \\ 
\midrule
Ganepola \& Wirasingha~\cite{ganepola2021automating} 
 & 2021 & Image generation, GAN compression; analysis of RL, EA, and~gradient-based strategies & Focuses on pre-2021 works, with~limited detail on mutation/operator design & \textls[-15]{Extends coverage to 2021–2025 with additional technical insights} \\ \midrule
Buthgamumudalige \& Wirasingha~\cite{buthgamumudalige2021neural} & 2021 & Transferability, supervised learning in NAS-GAN; evaluates IS and FID on CIFAR-10, STL-10 & Limited dataset diversity and minimal discussion of evolutionary operators & Expands evaluation to diverse datasets (CelebA, LSUN) and provides deeper operator-level~analysis \\ \midrule

Kang~et~al.~\cite{kang2023neural} & 2023 & NAS in computer vision tasks & Superficial NAS-GAN coverage & Provides in-depth NAS-GAN analysis \\\midrule

White~et~al.~\cite{white2023neural} & 2023 & Broad NAS survey (1000+ papers) & Minimal focus on GANs & Emphasizes GAN-specific~techniques \\ \midrule

Wang~et~al.~\cite{article} & 2024 & EC in GANs: architecture search, parameter tuning, loss function adaptation, and~synchronization strategies & Broad scope reduces NAS-specific depth; less focus on discrete architecture search & Goes beyond traditional EC-based methods , offering a holistic technical perspective and comprehensive analysis of diverse~approaches. \\ \midrule

This Review 
 & 2025 & Comprehensive NAS-GAN analysis & N/A & \textls[-25]{Synthesizes 2021--2025 works, addressing reproducibility, dataset diversity, and~technical gaps} \\ \bottomrule \end{tabularx} 
\end{adjustwidth}
\end{table}
\unskip

\subsection{Identified Gaps in the~Literature}
In our review of related studies, we identified a noticeable gap in the literature regarding the application of NAS to GANs. Given the rapid advancements in computer hardware capabilities and the increasing number of publications in the field, there is a compelling need for an up-to-date comprehensive review. Current surveys often focus on narrow aspects of NAS applied to GANs, such as specific datasets or image generation techniques. This paper aims to address these gaps by providing a broader analysis of NAS techniques and their applications to GANs, highlighting limitations and future research~opportunities.

To systematically differentiate our work from existing surveys, Table~\ref{tab:review_comparison} compares our review with two foundational prior studies by Sanepola \& Wirasingha (2021) \cite{ganepola2021automating} and Buthgamumudalige \& Wirasingha (2021) \cite{buthgamumudalige2021neural}.  

\begin{table}[H]
\small  
\caption{Comparative 
 Analysis of NAS-GAN~Reviews.}
\label{tab:review_comparison}

\begin{adjustwidth}{-\extralength}{0cm}
\begin{tabularx}{\fulllength}{LCCC}
\toprule
\multirow{2}{*}{\textbf{Aspect}} & 
\begin{tabular}[c]{@{}c@{}}\textbf{Ganepola}\\ \textbf{\& Wirasingha}\\ \textbf{(2021) \cite{ganepola2021automating}} \end{tabular} 
& \begin{tabular}[c]{@{}c@{}}\textbf{Buthgamumudalige}\\ \textbf{\& Wirasingha}\\ \textbf{(2021) \cite{buthgamumudalige2021neural}}\end{tabular} 
& \multirow{2}{*}{\textbf{This Review}} \\
\midrule
\vspace{6pt}{Search Strategies 
} & \begin{tabular}[c]{@{}c@{}}EA, RL,\\ Gradient-based\\\end{tabular} &\vspace{6pt}{RL, Gradient-based} & \begin{tabular}[c]{@{}c@{}}EA, RL,\\ Gradient-based\\\end{tabular} \\\midrule
\multirow{2}{*}{Datasets} & \multirow{2}{*}{CIFAR-10, STL-10} & \multirow{2}{*}{CIFAR-10, STL-10} & \begin{tabular}[c]{@{}c@{}}CIFAR-10, STL-10,\\ CelebA, LSUN,\\ MNIST\end{tabular} \\\midrule
\multirow{2}{*}{Evaluation Metrics} & \begin{tabular}[c]{@{}c@{}}IS, FID,\\ GPU days,\\ Search  space\end{tabular} 
 &\multirow{2}{*}{ IS, FID} & \begin{tabular}[c]{@{}c@{}}IS, FID,\\ Search spcae,\\ Computational Cost\end{tabular} \\\midrule
\multirow{2}{*}{Limitations Addressed} & \multirow{2}{*}{Focus on early NAS-GANs} & \multirow{2}{*}{Limited evaluation criteria} & \begin{tabular}[c]{@{}c@{}}Metric reliability,\\ Reproducibility,\\ Discriminator NAS gaps\end{tabular} \\\midrule
\vspace{6pt}Novel Contributions &\vspace{6pt} Initial NAS-GAN taxonomy & \vspace{6pt}Multi-criteria analysis & \begin{tabular}[c]{@{}c@{}}Analysis of recent NAS-GAN\\works,  dataset diversity\end{tabular} \\
\bottomrule
\end{tabularx}
\end{adjustwidth}
\end{table}

This comparison underscores our review’s unique focus on understudied challenges in NAS-GANs, such as reproducibility, while expanding the scope of datasets and evaluation criteria. Subsequent sections detail these contributions through quantitative and qualitative~analyses.

\section{Research~Methodology}\label{sec3}

{\textcolor{black}{
In this study, we present a comprehensive review of NAS techniques for GANs. We based our methodology on the guidelines proposed by Kitchenham~\cite{kitchenham2004procedures} and have applied strict quality assessment criteria. However, we note that this work is not a systematic review in the strict sense (e.g., following PRISMA guidelines) and does not fully adhere to all the reproducibility standards of a systematic review.
}

\subsection{Study Objectives and Research~Questions}

With the rapid advancement of AI and DL, GANs have emerged as powerful tools for generating realistic data across various domains. As~the complexity of GAN architectures grows, researchers have been increasingly interested in applying NAS to automate and optimize the design of GANs, allowing AI practitioners to focus on higher-level problems rather than manually crafting network architectures, which can be time-consuming and suboptimal~\cite{souza2021fine}. Thus, developing effective NAS methods for GANs could significantly enhance the quality and efficiency of generative~models.

The main objectives of this study are to review and analyze the state-of-the-art NAS approaches for GANs, their techniques, and~evaluation metrics. We conduct a comprehensive survey of the current landscape, focusing on the main types of NAS-GAN and identifying the important criterions of each type, such as underlying search space, techniques, search objectives, and~evaluation metrics. Moreover, this study aims to identify gaps in the existing literature and highlight promising future directions to further advance research in this~area.

For the purposes of this study, we developed comparison frameworks and applied them to a set of prominent studies. The~research questions that this review aims to answer are as follows:

\begin{itemize}
    \item RQ1: 
 What NAS approaches are applied to GANs in the literature?
    \begin{itemize}
        \item To address this question, we identified the approaches present in the literature, highlighting both their benefits and limitations. By~systematically reviewing existing methods, we were able to provide a comprehensive analysis that underscores the strengths of each approach while also acknowledging potential drawbacks and areas for improvement.
    \end{itemize}
    
    \item RQ2: What are the key search spaces explored in NAS-GAN?
    \begin{itemize}
        \item To address this research question, we examined the search spaces utilized in the studied approaches, exploring their applications and additional relevant aspects. This investigation provided insights into how different search spaces are leveraged within the context of the approaches, highlighting their effectiveness and areas for potential enhancement.
    \end{itemize}
    
    \item RQ3: What evaluation methods are used to assess the found architecture?
    \begin{itemize}
        \item To address this research question, we identified the metrics employed to evaluate NAS-GAN approaches and assessed their applicability. This analysis provided a detailed overview of the evaluation criteria used in the studies, examining their effectiveness in measuring the performance and suitability of the NAS-GAN approaches in various contexts.
    \end{itemize}
    
    \item RQ4: What are the gaps in the research on NAS in GANs?
    \begin{itemize}
        \item To address this research question, we studied and analyzed the relevant literature, providing a comprehensive review of existing studies and identifying areas for future research. This approach allowed us to suggest potential directions for future work based on the gaps and limitations identified in the current body of knowledge.
    \end{itemize}
\end{itemize}

\subsection{Search~Strategy}\label{sec3.2}

To support our study and answer our research questions, we utilized multiple information sources, focusing exclusively on scientific literature. We gathered relevant studies from key literature search engines and databases, including Google Scholar, ACM, ScienceDirect, IEEE, ArXiv, and~Springer. Additionally, to~enhance our search and uncover more pertinent studies that may not have appeared in our initial searches, we employed both backward and forward snowballing techniques. This comprehensive approach ensured that we captured a wide array of related studies, providing a robust foundation for our research analysis. The~annual 
distribution of the chosen literature is shown in Figure~\ref{fig:histogram}.
\begin{figure}[H]
\centering
\includegraphics[width=\textwidth]{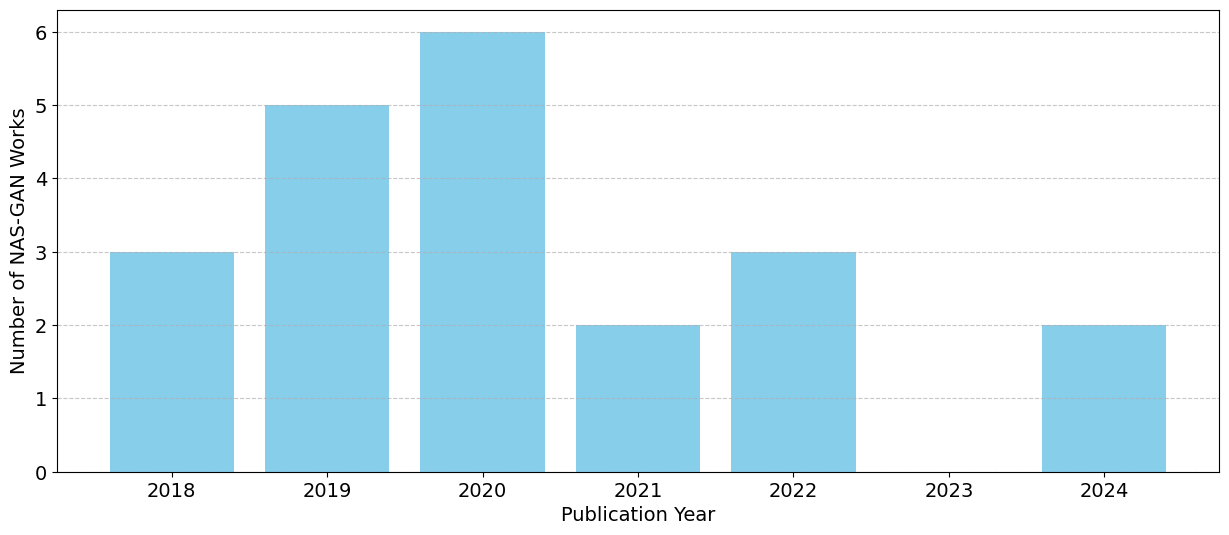} \caption{Literature~distribution.} \label{fig:histogram} 
\end{figure}

We employed the following search strings to identify relevant studies:

\begin{itemize}
    \item In Google Scholar, IEEE, ACM, and~Springer: (``Generative Adversarial Network'' OR ``GAN*'') AND (``Architecture'' OR ``Architectural'') AND (``Search'' OR ``Optimization'') AND (``Reinforcement Learning'' OR ``Policy'' OR ``Evolutionary'' OR ``Evolutionary Algorithm*'' OR ``Genetic Algorithm*'' OR ``Differential'' OR ``Gradient-based'')
    
    \item In Arxiv and Science Direct (we changed the search string because of limitations in their search tool): (``Generative Adversarial Network'' OR ``GAN'') AND (``Architecture'' OR ``Architectural'') AND (``Search'' OR ``Optimization'') AND (``Reinforcement Learning'' OR ``Policy'' OR ``Evolutionary'' OR ``Evolutionary Algorithm'' OR ``Genetic Algorithm'' OR ``Differential'' OR ``Gradient-based'').
\end{itemize}

\subsection{Study Selection and Quality~Assessment}

We selected studies through these steps:

\begin{enumerate}
    \item Initial selection: 
 We searched each database mentioned in Section 
 \ref{sec3.2}. We first chose studies based on their~titles.
    
    \item Filtering studies: To find the most relevant studies from our initial collection, we used our quality assessment criteria. We also looked at the abstract, introduction, and~conclusion of each study to filter them~further.
    
    \item Merging: After filtering, we had a group of studies relevant to our research. Some were duplicates because of overlaps in database results. We combined all the studies into one set, removing~duplicates.
    
    \item Snowballing: To find more related studies and make sure we didn't miss anything, we used backward and forward snowballing. Backward snowballing means looking at a study's reference list to find new papers~\cite{kitchenham2004procedures}. Forward snowballing means finding new papers that cite the study we're looking at~\cite{kitchenham2004procedures}. These processes helped us add new papers we didn't find in our first~search.
    
    \item Final Decision: After adding the new studies from snowballing, we filtered our set of studies one more time to get our final group of relevant studies.
\end{enumerate}

We applied a structured quality assessment using explicit inclusion and exclusion criteria to select the studies. The~criteria were defined as~follows:
\begin{enumerate}
    \item Inclusion Criteria:
    \begin{enumerate}
        \item Studies must be written in English.
        \item Studies must focus primarily on architecture search for GANs, providing clear details on the NAS techniques employed.
        \item Only peer-reviewed articles, conference papers, and~reputable preprints with sufficient methodological detail were considered.
    \end{enumerate}

    \item Exclusion Criteria:
    \begin{enumerate}
        \item Studies that focus primarily on topics other than architecture search (e.g., hyperparameter tuning, latent space exploration, or~unrelated GAN applications) were excluded.
        \item Studies lacking sufficient technical detail or methodological transparency regarding the NAS process were not considered.
        \item Duplicate studies identified across different databases were removed.
    \end{enumerate}
\end{enumerate}
\unskip

\subsection{Limitation on Research Methodology} \label{sec3.4}

Although a systematic review methodology would further enhance transparency and reproducibility, our study emphasizes comprehensiveness over strict adherence to systematic protocols. To~mitigate potential selection bias, we employed multiple literature databases along with backward and forward snowballing techniques, and~we applied explicit inclusion and exclusion criteria. Nevertheless, we acknowledge that the possibility of selection bias cannot be entirely excluded, and~future studies may benefit from a more formalized systematic review approach.

Additionally 
, while our review provides a qualitative synthesis of NAS-GAN methods, we have not performed a full statistical meta-analysis due to the heterogeneity in reported performance metrics (e.g., Inception Score, Fréchet Inception Distance, computational cost) across studies. We recognize that a quantitative meta-analysis could potentially offer deeper insights into the comparative effectiveness of these methods and recommend this as a direction for future research.

\subsection{Data Extraction and~Synthesis}
\label{sec3.5}
To ensure the integrity of extracted data and facilitate efficient management of the extraction process, we developed a structured comparative framework to address the research questions. This framework encompasses specific attributes for each research question, categorized according to their relevance. The~attributes are defined as follows:

\clearpage

\begin{itemize}
    \item RQ1: 

    \begin{itemize}
        \item Search Strategy: 
 This criterion examines the techniques and approaches employed in the solutions presented by relevant studies. These may include, but~are not limited to RL, EA, and~other methodologies prevalent in the existing literature.
        \item Search Type: This criterion identifies if the search strategy searches for both generator and discriminator networks, or~if it only searches for a generator network.
        \item Performance Assessment Strategy: This criterion examines how the search strategy estimates its current performance to guide the search.
        \item GPU Cost: This criterion identifies the search speed of the solutions presented by relevant studies based on its GPU usage.
        \item Advantages: This criterion examines the positive outcomes and potential merits associated with each methodological approach discussed in the existing literature.
        \item Disadvantages: This criterion identifies and analyzes any constraints, shortcomings, or~negative aspects (where applicable) of the investigated approach.
    \end{itemize}
    
    \item RQ2:
    \begin{itemize}
        \item Search Space: This criterion examines the type of search space used to encode the network component in the solutions presented by relevant studies. These may include, but~are not limited to, Cell-based or Entire/Chain-Structured and other methodologies prevalent in the existing literature.
    \end{itemize}
    
    \item RQ3:
    \begin{itemize}
        \item Evaluation Metrics: This criterion examines evaluation measurements used to evaluate the performance of the solution presented by the relevant studies.
        \item Dataset: This criterion identifies the datasets used in the solution presented by relevant studies.
        \item Supported Generation Type: This criterion identifies the types of generation tasks that are supported by the solution approaches. These may include, unconditional image generation, conditional images generation or both
    \end{itemize}
\end{itemize}

\textls[-15]{The findings derived from the literature review were critically examined in Section~\ref{sec4}, aligning with the established research questions and their corresponding comparative~criteria.}

\section{Results and~Discussion}\label{sec4}

Our analysis of the collected research yielded several important discoveries and observations. In~the following section, we present these findings and provide answers to our research questions based on our comprehensive review of the~literature.

\subsection{RQ1: What NAS Approaches Are Applied to GANs in the Literature?}

Our literature review uncovered a variety of strategies for implementing NAS-GAN approaches. We grouped the studies based on their core algorithms and common traits. This resulted in three main categories: Evolutionary Algorithms, Reinforcement Learning, and~Gradient-based Algorithms. As~shown in Figure~\ref{fig:nasgan_pie_chart}, Evolutionary Algorithms dominate the literature with 11 works, while both Reinforcement Learning and Gradient-Based Approaches each account for 5 works. We explore each category in depth, detailing the specific approaches and discussing their respective strengths and~limitations.

\begin{figure}[H]
  \hspace{-9pt}\includegraphics[width=0.5\textwidth]{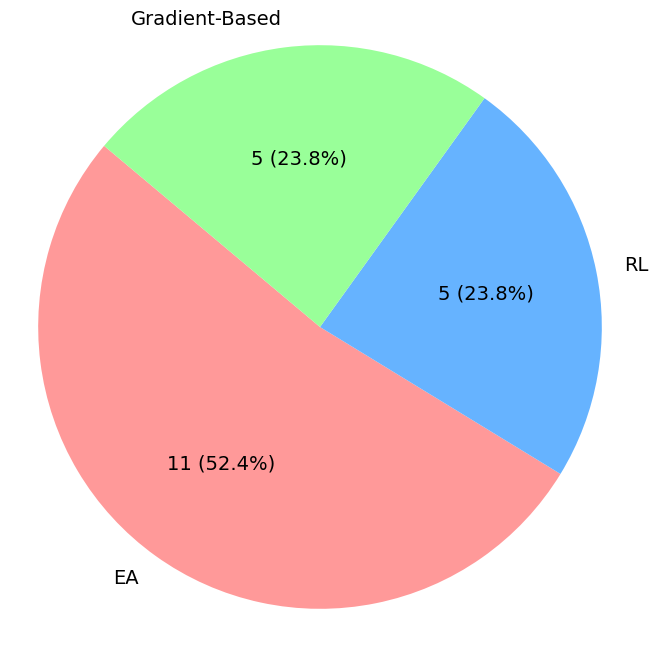}
  \caption{Distribution of NAS-GAN Methods by Search~Strategy.}
  \label{fig:nasgan_pie_chart}
\end{figure}
\unskip

\subsubsection{Evolutionary Algorithms~Approaches}

EA is a family of metaheuristic optimization algorithms, inspired by collective behaviors observed in nature~\cite{talbi2009metaheuristics}. These algorithms can navigate complex high-dimensional search spaces to find near-optimal solutions, making them well-suited for automatically finding answers to the many interdependent architectural decisions in deep \mbox{neural~networks.}

EAs have been increasingly applied to the field of GAN architecture search, with~several notable approaches emerging in recent years. Wang~et~al. (2019) \cite{wang2019evolutionary} proposed E-GAN, which evolves a population of generators using different adversarial objectives as mutation operations. Specifically, they employed three different objectives: Minimax objective, Heuristic objective Least-squares objective. This method adapts to the discriminator and overcomes limitations of individual adversarial objectives, requiring 1.25 GPU Days for searching. Yet only the top-performing generator persists, which might limit diversity. Al-dujaili~et~al. (2018) \cite{al2018towards} introduced Lipizzaner, a~coevolutionary framework that evolves the internal parameters (weights) of fixed generator and discriminator architectures. Building on this, Toutouh~et~al. (2019) \cite{toutouh2019spatial} developed Mustangs, a~hybrid approach combining elements of E-GAN and Lipizzaner, incorporating multiple loss functions as mutation operators and using a spatial grid coevolution~scheme.

Complementing these approaches, Garciarena~et~al. (2018) \cite{10.1145/3205455.3205550} introduced EvoGAN a neuro-evolutionary framework that evolves GAN architectures by encoding both the network topology (e.g., number of hidden layers, activation functions, weight initialization) and key training parameters such as the loss function and synchronization frequency between the generator and discriminator. Their method leverages a genetic algorithm with specialized mutation operators to navigate a flexible search space, employing Pareto set approximation and the inverted generational distance (IGD) as a benchmark for evaluating mode collapse and diversity. Notably, EvoGAN scales to 784 variables, though~retaining only the top-performing architecture per generation might limit diversity. The~framework was evaluated on a set of bi‐objective continuous test problems for Pareto set approximation. Performance was measured using the inverted generational distance (IGD); for example, on~function F1, IGD was reduced by up to two orders of magnitude in both low- (n = 10) and high-dimensional (n = 784) settings, with~additional tests demonstrating transferability across~functions.

Similarly, Lu~et~al. (2018) \cite{lu2018autonomously} propose a GA-assisted Bi-GAN framework that autonomously refines deep neural network parameters by combining discrete GA-based decisions with continuous refinement via a bi-generative adversarial network. This approach optimizes parameters—including the number of neurons, filters, layers, and~even the inclusion of dropout or pooling layers—thereby enabling the network to self-configure its architecture and hyper-parameters during training. By~bridging exploration and exploitation, this framework overcomes the limitations of fixed discrete candidate sets typically associated with GA-only approaches. Bi-GAN was tested on the voxelized ModelNet40 dataset~\cite{7298801} (3D CAD models of 40 object classes). Their experiments showed that the proposed method achieved an accuracy of 85.20\%, outperforming the baseline 3D Shapenet (84.17\%) and two GA-only approaches (small-set GA: 82.94\%; large-set GA: 36.41\%).

In other studies, multi-objective function and multi-stage search have been utilized. Du~et~al. (2020) \cite{du2020structure}, using the Non-Dominated Sorting Genetic Algorithm II (NSGA-II) algorithm for multi-objective optimization of DCGAN structures (NSGA-II DCGAN), specifically used True Positive Rate (1-TPR) and False Positive Rate (FPR). Implementing two-stage searches, Lin~et~al. (2022) \cite{lin2022evolutionary} proposed EAS-GAN, an~evolutionary architectural search method that searches for generator using a multi-objective function. Generator architectures evolve using three objective functions as mutation operators: minimax, least-squares, and~hinge loss, and~then a traditional adversarial training for the discriminator weights. The~search process takes 1 GPU Day. Costa~et~al. (2019) \cite{costa2019coegan} developed COEGAN, a~neuroevolutionary and coevolutionary approach that evolves both generator and discriminator architectures. Utilizing different objectives for the generator and discriminator populations, the~discriminator objective is the adversarial loss while the generator objective is the FID~score.

Ying~et~al. (2022) \cite{ying2022eagan} introduced EAGAN, a~two-stage evolutionary NAS that decouples the search for generator and discriminator networks. It uses Multi-objective Pareto optimization, considering model size, IS, and~FID as objectives. It efficiently searches for the optimal GAN in 1.2 GPU Days. Finally, Xue et.al. (2024) \cite{xue2024evolutionary} introduced EWSGAN, the~method employs a two-step process: first training a super net generator using weight sharing and single-path sampling, then utilizing NSGA-II to search for optimal subnets. EWSGAN focuses solely on searching for generator architectures while using a fixed discriminator. The~search process is highly efficient, completing in just 1 GPU Day. It has a multi-objective function, simultaneously optimizing IS and FID. The~approach offers several advantages, including an efficient search process due to weight sharing and low-fidelity evaluation, improved stability through fair single-path sampling and a commonality-based discarding strategy. However, a~potential drawback is the discarding strategy that may require further investigation, and~possible challenges in scaling to higher-resolution~datasets.

Recent work has explored conducting hyperparameter optimization (HPO) and NAS jointly. Kobayashi and Nagao (2020) \cite{kobayashi2020multi} proposed searching the architecture and hyperparameters of GANs using multi-objective evolutionary algorithms. They used Cartesian Genetic Programming (CGP) with NSGA-II (NSGA-II with CPG) to evolve generator and discriminator architectures simultaneously, optimizing various hyperparameters. The~multi-objective fitness function maximized IS score and minimized FID. It is considered one of the first works that implemented HPO and NAS for GANs. However, a~limitation of the work is the restriction of the network size, limiting the scalability of the method. Also, although~training time was not reported, the~authors stated that the search efficiency needs to be~improved.

Common benefits across these approaches include improved generative performance, increased architectural efficiency, and~enhanced adaptability to different datasets. The~use of multi-objective optimization in several approaches allowed for balancing multiple performance criteria simultaneously, leading to more robust solutions. Despite that, EA approaches' performance tends to rely heavily on the objective function. Another limitation to EA approaches is the search space formulation: EA approaches were mainly used with discrete search~space.

\subsubsection{Reinforcement Learning~Approaches}

RL is a paradigm of machine learning inspired by behavioral psychology, where an agent learns to make decisions by interacting with an environment~\cite{sutton2018reinforcement}. This approach enables systems to learn optimal policies in complex, dynamic settings by maximizing cumulative rewards, making it well-suited for solving sequential decision-making problems in searching for neural network architecture. RL approaches can be categorized as on-policy and off-policy~approaches.
\begin{itemize}
\item {On-Policy Search 
} 
\end{itemize}
Gong~et~al. (2019) \cite{gong2019autogan} introduced AutoGAN, a~novel method that uses a Recurrent Neural Network (RNN) controller to guide the search process for generator architectures. AutoGAN employs a multi-level search strategy with beam search, using SoftMax predictions for sampling architecture variations. While it focuses solely on optimizing the generator, with~the discriminator growing in a pre-defined manner, AutoGAN demonstrates the potential of using IS as a reward signal in RL for GAN design, in~just 2 GPU Days. However, this approach is limited by its lack of discriminator architecture search and potential scalability issues for higher resolution images. 
Wang~et~al. (2019) \cite{wang2019agan} developed AGAN, a~reinforcement learning framework that simultaneously searches both generator and discriminator architectures. AGAN employs a two-layer LSTM controller RNN and uses policy gradient with REINFORCE, incorporating an entropy bonus for exploration. This approach stands out by allowing for arbitrary cell topologies and demonstrating adaptability to different image sizes and complexities. AGAN uses a shaped reward function based on IS, potentially offering more nuanced guidance for architecture optimization. However, the~simultaneous search of both generator and discriminator architectures lead to increased computational requirements as it takes 1200 GPU Days and necessitates careful tuning of the reward function~shaping.

Zhou~et~al. (2020) \cite{9839586} proposed Multi-Net NAS (MN-NAS), a~novel approach that leverages reinforcement learning to design class-aware generators for conditional GANs (cGANs). MN-NAS employs an MDP with a moving average mechanism to sample and evaluate candidate architectures. The~key innovation lies in its ability to search for distinct generator architectures for each class within a single search procedure, addressing the challenge of combinatorial explosion as the number of classes increases. The~search space includes regular convolutions and class-modulated convolutions (CMconv), which allow for the sharing of training data across different architectures, mitigating the issue of insufficient data per class. MN-NAS also introduces mixed-architecture optimization, enabling efficient parallelization of the search and re-training processes. The~method demonstrates competitive performance on CIFAR10 and CIFAR100. However, the~approach is limited by its focus on generator architecture search, with~the discriminator following a predefined design, and~the potential scalability issues when applied to datasets with a very large number of~classes.
\begin{itemize}
\item{Off-Policy Search} 
\end{itemize}
Tian~et~al. (2020) \cite{tian2020off} proposed E$^2$GAN, an~off-policy reinforcement learning framework that reformulates the search problem as a Markov Decision Process (MDP). E$^2$GAN utilizes a soft actor-critic algorithm and a progressive state representation, significantly improving sample efficiency. The~framework implements exploration and exploitation periods, focusing on cell architectures. E$^2$GAN's key innovation lies in its ability to discover competitive GAN architectures in just 7 GPU Hours, using a combined reward function of IS and FID. While E$^2$GAN offers rapid architecture discovery, a~limitation of this approach is that it only searches for the~generator.

Li et al. (2022) \cite{9699403} introduced T2IGAN the first work to apply NAS principles for designing GANs that integrate transformer modules into the text-to-image synthesis framework. In~T2IGAN, the~architecture search is formulated as a MDP, and~an RL-based search strategy is adopted to efficiently navigate a cell-based search space that encompasses both convolutional operations and lightweight transformer components. The~search processjointly optimizes both the cell structure and the associated operation weights using a composite reward function based on metrics such as IS and FID. Ultimately, the~final generator is constructed by stacking the best-performing cells discovered during the search, yielding a competitive architecture for text-to-image synthesis. By~leveraging an off-policy RL, T2IGAN is able to significantly reduce the computational burden, achieving competitive performance in a fraction of the search time compared to earlier approaches. While the method primarily focuses on optimizing the generator architecture (with the discriminator following a pre-defined design), it effectively demonstrates the feasibility and advantages of combining transformer-based representations with adversarial learning. Overall, T2IGAN represents a meaningful step toward automated, efficient GAN design and underscores the potential of off-policy RL techniques in advancing generative model architecture~search.

These studies collectively showcase the evolution of reinforcement learning applications in GAN architecture search. From~AutoGAN's focus on generator optimization, through AGAN's comprehensive search of both GAN components, to~off-policy frameworks like E$^2$GAN and T2IGAN that emphasize efficiency improvements, each approach contributes unique strengths to the field. However, a~notable challenge of these RL-based methods is the considerable variation in training time, which can complicate scalability and practical~deployment. 

\subsubsection{Gradient-Based~Approaches}

Gradient-based algorithms are a fundamental class of optimization methods~\cite{goodfellow2016deep}. These techniques leverage the gradient of an objective function to iteratively update model parameters, efficiently navigating high-dimensional parameter spaces to minimize loss or maximize performance. Gradient-based algorithms are not typically used directly for searching neural network architectures. However, there are related approaches that use gradients to help with architecture~search.

Gradient-based NAS methods have been successfully applied to optimize GAN architectures. AdversarialNAS, proposed by Gao~et~al. (2020) \cite{gao2020adversarialnas}, uses a differentiable approach to simultaneously search for both generator and discriminator architectures. It employs an adversarial search mechanism and can discover high-performing architectures in just 1 GPU Day. The~method boasts a large search space and demonstrates good transferability and~scalability. 

Doveh and Giryes (2021) \cite{doveh2021degas} introduced DEGAS, a~gradient-based method focusing solely on generator architecture search. DEGAS reformulates the problem as a differentiable optimization task and utilizes the Global Latent Optimization technique to avoid adversarial training instabilities during search. It can find competitive generator architectures in 1.16 GPU Days, significantly faster than most previous reinforcement learning-based methods. A~limitation of DEGAS is that it only searches for Generator architectures. Searching for Discriminator architectures may have an impact on~results. 

GAN Compression, introduced by Li~et~al. (2020) \cite{Li_2020_CVPR}, presents a gradient‐based framework for compressing the generator component of conditional GANs. Rather than relying on reinforcement learning or evolutionary methods, this approach decouples network training from architecture search by first training a “once‐for‐all” generator via standard gradient descent with weight sharing. In~this phase, the~network is designed to support a vast array of sub‑networks through flexible channel configurations. Once trained, the~method efficiently evaluates these sub‑networks using differentiable losses—including intermediate feature distillation—to select the most efficient architecture that satisfies a specified computational budget. This gradient‐based sub‑network selection allows GAN Compression to dramatically reduce both the inference time and model size; empirical results show reductions in MACs by up to 21× on models such as CycleGAN, pix2pix, and~GauGAN while preserving high visual fidelity. A~key strength of GAN Compression is its model‐agnostic nature and its ability to stabilize GAN training via knowledge transfer, thereby enabling interactive conditional GAN applications on resource‐constrained~devices.

Tian~et~al. (2021) \cite{tian2021alphagan} proposed alphaGAN, a~fully differentiable architecture search framework that formulates the problem as a bi-level minimax optimization. A~key innovation is the use of the duality gap as a differentiable evaluation metric. alphaGAN can efficiently discover high-performing architectures for both conventional GANs and StyleGAN2 in just 3 GPU Hours, demonstrating strong transferability and~scalability.

Xue~et~al. (2024) \cite{10477508} presented Differentiable architecture search with attention mechanisms for GANs (DAMGAN)  an innovative evolution in gradient-based NAS approaches. Unlike traditional methods that rely on fixed architectural parameters, DAMGAN leverages a dual-attention strategy to guide the search process, thereby enhancing training stability and efficiency. In~this approach, a~generator supernet is constructed with two distinct attention mechanisms: up-attention (UA) and down-attention (DA). UA is used to select the most salient feature maps before candidate operations are applied, effectively reducing computational overhead and ensuring that only the most informative features contribute to the network’s evolution. In~parallel, DA evaluates the outputs of multiple candidate operations to assign importance weights, replacing the conventional reliance on architectural parameters. This refined mechanism allows DAMGAN to efficiently determine the optimal candidate operations and construct a high-performing subnet with a remarkably low computational cost. Empirical results demonstrate that DAMGAN achieves competitive performance on benchmark datasets such as CIFAR-10—with while completing the search in only 0.09 GPU days. The~method’s scalability and transferability are further validated by its successful application to larger datasets like STL-10 and~CelebA. 

\subsection{RQ2: What Are the Key Search Spaces Explored in NAS-GAN?}

The search space defines which neural architectures can be represented and potentially discovered by a NAS method. The~design of the search space is crucial as it incorporates prior knowledge about architectures that are likely to perform well on a given task, while also influencing the difficulty of the optimization problem. Among~the search spaces of NAS in GANs literature, the~cell-based structure is the most used type, and~the entire/chain-structure search space has also been employed. As~illustrated in Figure~\ref{fig:search_space_pie_chart}, cell-based approaches account for the majority of the works, followed by chain-based, hybrid, and~custom~approaches.

\begin{figure}[H]
 \hspace{-9pt} \includegraphics[width=0.5\textwidth]{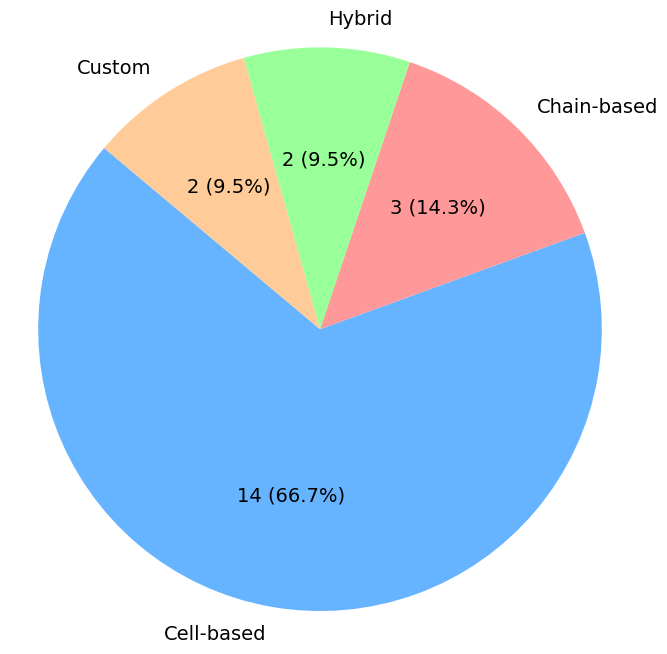}
  \caption{Distribution of Search Space Types in NAS-GAN~Literature.}
  \label{fig:search_space_pie_chart}
\end{figure}
\unskip

\subsubsection{Entire/Chain-Structure~Space}

Entire/chain-structure neural networks represent one of the simplest search spaces. In~this space, an~architecture can be described as a sequence of multiple layers, where the output of a layer serves as input for the next~layer.

NSGA-II DCGAN~\cite{du2020structure} presents an entire/chain -structure search space. This search space encompasses various architectural elements for both the generator and discriminator networks. For~the generator, the~search space includes the number of convolutional layers, the~number of filters (kernels) in each layer, kernel sizes, padding settings, and~activation functions. The~discriminator's search space mirrors that of the generator, allowing for a symmetric exploration of network architectures. The~framework employs a flexible approach, treating the number of layers as a significant parameter, thus allowing for variable-depth networks. This flexibility extends to other hyperparameters of the network structure, creating a multidimensional search space. The~structural parameters are encoded as solution vectors, forming a continuous search~domain.

COEGAN~\cite{costa2019coegan} employs a chain-structured search space to evolve both the Generator and Discriminator networks in a GAN. This search space incorporates three types of layers: linear, convolution, and~transpose convolution, each of which can have an activation function randomly selected from options like ReLU, LeakyReLU, ELU, Sigmoid, and~Tanh. The~convolution and transpose convolution layers vary in the number of output channels, while linear layers vary in the number of output features. The~architecture evolves through mutations that can add, remove, or~modify layers, with~the number of layers capped at six in the experiments, although~this parameter is adjustable. The~search space supports dynamic calculation of parameters such as stride and kernel size for convolutional layers, based on the required output size of each layer. Designed to incrementally increase network complexity over generations, the~search space is structured to safeguard novel architectures through a speciation~mechanism.

DEGAS~\cite{doveh2021degas} uses an entire/chain-structure search space, focusing on searching neural architectures for the Generator network efficiently. It searches for the whole network architecture globally, divided into three parts: a fixed first part with linear operation and reshape, a~searchable middle part, and~a fixed last part with batch normalization, ReLU, convolution and Tanh. The~searchable part includes normal operations that maintain input size and up-sample operations that increase it. Normal operations encompass various combinations of batch norm, ReLU, convolutions, pooling, skip connections, and~dilated convolutions. Up-sample operations include deconvolutions and nearest neighbor upsampling with convolutions. Connections between feature maps use Mixed Operations with these defined~operations.

The entire/chain-structure approach offers more freedom in architecture design but may lead to a larger search space, potentially increasing computational costs. DEGAS's approach of having fixed parts could help reduce the search space while still allowing for significant architectural~exploration.

\subsubsection{Cell-Based~Space}

Among the search spaces of NAS in GANs literature, the~cell-based structure is the most used type, instead of searching for entire architectures, this method focuses on finding smaller architectural building blocks called cells or motifs. The~final architecture is then created by stacking these cells in a predefined manner. It significantly reduces the size of the search space, as~cells usually consist of fewer layers than complete architectures. This can lead to substantial speed-ups in the search process. When using a cell-based search space, a~new design choice arises: how to choose the macro-architecture, i.e.,~how many cells to use and how to connect them to build the actual~model.

Lipizzaner~\cite{al2018towards} employs a cell-based search space, distributing Generators and Discriminators across a 2D toroidal grid. Each cell contains one or more GANs, interacting within defined neighborhoods (typically a center cell and its four adjacent cells). The~search space encompasses neural network parameters, hyperparameters, and~mixture weights. This structure allows for local adaptation and diversity maintenance, supporting various neural architectures (e.g., multi-layer perceptrons or deconvolutional GANs) within cells. Mustangs~\cite{toutouh2019spatial} adopts the search space to~Lipizzaner.

AutoGAN's~\cite{gong2019autogan} search space is designed to identify optimal Generator network architectures. It encompasses five key elements: skip connections, convolutional block varieties, normalization block options, upsampling techniques, and~an in-cell shortcut indicator. The~convolutional block category includes both pre- and post-activation variants, while the normalization block offers batch, instance, and~no normalization choices. Upsampling options consist of bi-linear, nearest-neighbor, and~stride-two deconvolution methods. To~facilitate a direct comparison of search strategy efficacy and efficiency, E$^2$GAN~\cite{tian2020off} and EWSGAN~\cite{xue2024evolutionary} adopt a search space identical to~AutoGAN's.

AGAN's~\cite{wang2019agan} search space incorporates architectural design principles for both the Generator and Discriminator. The~search methodology employs reinforcement learning, utilizing a controller comprised of a two-layer LSTM network. This controller navigates the search space to identify high-performing architectures within the predefined parameters. For~upsampling, the~system employs transposed convolution and nearest-neighbor interpolation techniques. The~downsampling modules are constructed using two distinct atomic operations: one applies convolution before stride-2 average pooling, while the other reverses this order, performing stride-2 average pooling followed by convolution.
AlphaGAN's~\cite{tian2021alphagan} search space is composed of two main categories: normal operations and upsampling operations. Normal operations primarily consist of convolutional blocks. For~upsampling, the~framework employs three distinct methods: deconvolution, nearest-neighbor interpolation, and~bi-linear~interpolation.

AdversarialNAS~\cite{gao2020adversarialnas} introduced an extensive search space $10^3$ for GANs, resulting in a continuous search domain. This approach employs probability distributions, with~architecture representation defined by a set of continuous variables. The~framework's search space is cell-based. For~the Generator, it includes a variety of normal operations and upsampling techniques, such as bilinear interpolation, nearest-neighbor interpolation, and~transposed convolution. The~Discriminator's structure incorporates both normal operations and downsampling methods, including max pooling, average pooling, and~convolutions. This expansive and flexible search space allows for a comprehensive exploration of potential architectures, enhancing the adaptability and optimization of the GANs developed through this framework. EAGAN~\cite{ying2022eagan} implements the same search space to allow fair~comparisons.

EAS-GAN's~\cite{lin2022evolutionary} search space focuses on the Generator's architecture using a cell-based approach. The~search employs an evolutionary algorithm, treating the generator as an evolving supernet composed of multiple Directed Acyclic Graphs (DAGs). Each cell is represented by a DAG with N nodes, where edges between nodes are candidate operations including various convolutions (1 $\times$ 
 1, 3 $\times$ 3, 5 $\times$ 5), dilated convolutions, skip-connections, and~zero operations. Upsampling options include transposed convolution 3 $\times$ 3, nearest neighbor, and~bilinear interpolation. The~evolutionary process optimizes both cell structure and operation weights simultaneously. Architectures are evaluated against a Discriminator serving as the evolutionary environment. The~final generator is constructed by stacking the best-performing discovered~cells.

T2IGAN's~\cite{9699403} search space is designed to optimize the generator architecture via a cell‐based approach. The~generator is modeled as a supernet comprised of multiple cells, each represented by a DAG with a fixed number of nodes (typically 4). In~each cell, edges correspond to candidate operations drawn from two primary categories: conventional convolutional operations and transformer modules. The~convolutional candidates include operations with kernel sizes of 3 × 3 and 5 × 5, while the transformer candidates incorporate multi‐head self‐attention (configured with 4 or 8 heads) followed by a position-wise feed-forward network with an expansion factor of 4 and an internal dimension of 128. Additionally, a~zero operation is provided, allowing the cell to bypass an edge when beneficial. Upsampling choices within the cell-based framework include nearest neighbor and bilinear interpolation techniques to progressively scale feature maps from an initial low resolution (e.g., 4 × 4) to higher~resolutions. 

MN-NAS~\cite{9839586} propose a search space tailored for designing class-aware generators in conditional GANs (cGANs). The~search space follows a cell-based structure, where the generator is composed of multiple cells, each containing a fixed number of nodes (e.g., 4~nodes). Each cell is designed to maintain the spatial resolution and channel dimensions of the input data, ensuring consistency throughout the network. Within~each cell, edges between nodes represent candidate operations, which are selected from a set of operators that include regular convolutions (RConv) and class-modulated convolutions (CMConv). The~RConv operator performs standard convolutional operations, while the CMConv operator introduces class-specific information by modulating the convolutional weights using a class-conditional vector. This modulation is achieved through a combination of affine transformations and normalization steps, allowing the network to share convolutional weights across different classes while still incorporating class-specific adjustments. The~search space also includes a zero operation, which allows the network to skip certain edges when necessary, providing flexibility in architecture design. The~overall architecture is constructed by stacking multiple cells, with~each cell contributing to the progressive upsampling of the input latent vector to generate high-resolution~images. 

GAN Compression’s~\cite{Li_2020_CVPR} search space differs from the typical cell-based approach. Instead of designing entirely new cells or motifs, GAN Compression leverages the structure of a pre-trained teacher generator and focuses on automatically reducing its redundancy by searching over channel configurations. In~this framework, a~“once-for-all” network is first trained via standard gradient descent with weight sharing; this super-network supports a wide range of sub-networks, each corresponding to a unique assignment of channel widths across the generator’s layers. Each convolutional layer is allowed to choose its number of channels from a discrete set (typically multiples of 8), which reflects a trade-off between computational efficiency and hardware parallelism. The~overall search space is defined as the combinatorial product of the candidate channel numbers for all layers, yielding a large—but highly structured—domain. This formulation enables fine-grained architectural optimization: the method can automatically determine which layers are more amenable to aggressive channel reduction without significantly degrading~performance.

In a similar vein, DAMGAN’s~\cite{10477508} search space is crafted to facilitate the efficient discovery of high‑performance generator architectures through a differentiable approach enhanced by attention mechanisms. The~search space is built upon a generator supernet organized into a series of cells, each containing multiple interconnected nodes. For~each pair of nodes, the~candidate operations are divided into two distinct groups. In~connections emanating from the input node, the~search space includes upsampling operations such as nearest‑neighbor sampling, bilinear interpolation sampling, and~transposed convolution. For~all other inter‑node connections, the~search space comprises convolutional operations, including standard convolutions with kernel sizes of 1 × 1, 3 × 3, and~5 × 5, as~well as depthwise separable convolutions with kernel sizes of 3 × 3, 5 × 5, and~7 × 7. Uniquely, DAMGAN integrates two attention mechanisms—up‑attention (UA) and down‑attention (DA)—between each pair of nodes. UA selectively filters and forwards the most salient feature maps to the candidate operations, while DA evaluates and assigns importance weights to the outputs of these operations. This dual‑attention strategy not only refines the selection process by effectively mapping the significance of each operation but also broadens the search space to encompass the dynamic interplay between feature selection and operation efficacy, ultimately leading to more robust and computationally efficient architecture~discovery.

The cell-based approach seems to be favored in NAS-GAN research, likely due to its ability to reduce the search space size while still allowing for complex architectures. This approach can lead to more efficient searches and potentially better~scalability.

In addition to the above-discussed search spaces, hybrid approaches have also been explored. EvoGAN~\cite{10.1145/3205455.3205550}  defines a comprehensive search space that spans both architectural and training parameters. In~this framework, the~entire network specification—including the generator and discriminator topologies—is encoded as lists that capture discrete decisions (e.g., number of hidden layers, activation functions, weight initialization methods) as well as continuous parameters such as loss functions and update (synchronization) frequencies. Specialized mutation operators (e.g., layer\_change, activ\_change, latent\_change) enable the GA to explore this vast and flexible space, which scales up to 784 variables and permits the discovery of transferable~architectures.

Similarly, the~Bi-GAN framework~\cite{lu2018autonomously} partitions the search space into discrete and continuous components. The~discrete component governs fixed architectural decisions—such as the number of layers, the~choice of activation functions, and~binary options like dropout, batch normalization, and~pooling—while the continuous component, refined via the Bi-GAN, optimizes parameter values such as the number of neurons in fully-connected layers and the number of filters in convolutional layers. This hybrid design enables a more nuanced exploration of the architectural design space by balancing rigid candidate sets with flexible parameter~tuning.

One key limitation is the lack of clarity in reporting search space sizes. Many papers fail to provide detailed implementation information, making it difficult to calculate or infer the exact dimensions of the search space. This ambiguity hinders direct comparisons between different approaches and impedes our understanding of the relative efficiency of various methods. Moving forward, researchers should strive for more transparent and comprehensive reporting of search space sizes and implementation~details.

There is also an apparent imbalance in the focus of current research, with~many approaches prioritizing the optimization of Generator architectures over Discriminators. While the Generator plays a crucial role in GAN performance, the~Discriminator is equally important. The~diversity of search strategies employed in the field suggests that the optimal approach may vary depending on the specific problem or~dataset.

As search spaces continue to grow in size and complexity, computational efficiency becomes increasingly critical. Innovative approaches that balance the trade-off between search space size and flexibility, such as DEGAS~\cite{doveh2021degas} combining fixed and searchable architectural components, could prove valuable. Lastly, the~potential for transferability of discovered architectures across different tasks or datasets is an area that warrants further investigation. While cell-based approaches might inherently offer better transferability, this aspect isn't explicitly addressed in much of the current literature. Table~\ref{tab:gan-comparison} presents a summary of reviewed~literature.
\begin{table}[H]
\small
\caption{Comparison of different GAN architecture search~methods.}
\label{tab:gan-comparison}

\begin{adjustwidth}{-\extralength}{0cm}
\begin{tabularx}{\fulllength}{lcccccccc}
\toprule
\textbf{Method} & 
\begin{tabular}[c]{@{}c@{}}\textbf{Search}\\\textbf{Strategy}\end{tabular} & 
\begin{tabular}[c]{@{}c@{}}\textbf{Searched}\\\textbf{Network}\end{tabular} & 
\begin{tabular}[c]{@{}c@{}}\textbf{Architecture} \\  \textbf{Modification}\\\textbf{Technique}\end{tabular} & 
\begin{tabular}[c]{@{}c@{}}\textbf{Optimization} \\ \textbf{Objective}\end{tabular} & 
\begin{tabular}[c]{@{}c@{}}\textbf{GPU}\\\textbf{Type}\end{tabular} & 
\begin{tabular}[c]{@{}c@{}}\textbf{GPU}\\\textbf{Days}\end{tabular} & 
\begin{tabular}[c]{@{}c@{}}\textbf{Search}\\\textbf{Space}\\\textbf{Type}\end{tabular} & 
\begin{tabular}[c]{@{}c@{}}\textbf{Search}\\\textbf{Space}\\\textbf{Size}\end{tabular} \\
\midrule
AutoGAN  & RL & G Only & \begin{tabular}[c]{@{}c@{}}RNN\\Controller\end{tabular} & IS & 2080Ti & 2 & Cell & $10^5$ \\
AGAN  & RL & G and D & \begin{tabular}[c]{@{}c@{}}RNN\\Controller\end{tabular} & IS & Titan-X & 1200 & Cell & $10^5$ \\
MN-NAS & RL & G Only & MDP & IS & 1080Ti & - & Cell & $10^{27}$ \\
E$^2$GAN & RL & G Only & MDP & IS \& FID & 2080Ti & 0.3 & Cell & - \\
T2IGAN & RL & G Only & MDP & \begin{tabular}[c]{@{}c@{}}IS \& FID\end{tabular} & V100 & 0.42 & Cell & $10^{25}$ \\
\midrule
AdversarialNAS  & Gradient & G and D & - & GAN objective & \begin{tabular}[c]{@{}c@{}}2 X\\2080Ti\end{tabular} & 1 & Cell & $10^{38}$ \\
DEGAS  & Gradient & G Only & - & Reconstruction loss & Titan-X & 4 & Chain & $10^8$ \\
GAN Compression & Gradient & G Only & - & \begin{tabular}[c]{@{}c@{}}GAN Objective + \\ reconstruction loss\end{tabular} & 2080Ti & - & Cell & $10^{9}$ \\
alphaGAN & Gradient & G Only & - & \begin{tabular}[c]{@{}c@{}}Duality Gap Loss\\\& GAN Objective\end{tabular} & \begin{tabular}[c]{@{}c@{}}Tesla\\P40\end{tabular} & 0.15 & Cell & $10^{11}$ \\
DAMGAN & Gradient & G Only & \begin{tabular}[c]{@{}c@{}}Dual-Attention \\ Mechanisms\end{tabular} & GAN objective & 3090 & 0.09 & Cell & –\\
\midrule
EvoGAN & EA & G and D & Mutation  & Custom & NA & - & Hybrid & - \\ Bi-GAN & EA & G Only & \begin{tabular}[c]{@{}c@{}}Mutation \&\\ Continuous \\ Refinement\end{tabular} & Accuracy & NA & - & Hybrid & - \\
NSGA-II DCGAN  & EA & G Only & \begin{tabular}[c]{@{}c@{}}Crossover \& \\ Mutation\end{tabular} & Custom & - & - & Chain & - \\
E-GAN  & EA & G and D & Mutation & Custom & 1080TI & 1.25 & - & - \\
Lipizzaner  & EA & G and D & Mutation & GAN objective & - & - & Cell & - \\
Mustangs  & EA & G and D & Mutation & GAN objective & - & - & Cell & - \\
COEGAN & EA & G and D & Mutation & \begin{tabular}[c]{@{}c@{}}FID \& \\ Discriminator loss\end{tabular} & - & - & Chain & - \\
EAGAN  & EA & G and D & \begin{tabular}[c]{@{}c@{}}Crossover \& \\ Mutation\end{tabular} & IS - FID & NA & 1.2 & Cell & $10^{38}$ \\
EAS-GAN  & EA & G Only & Mutation & Custom & 3090 & 1 & Cell & - \\
EWSGAN  & EA & G Only & \begin{tabular}[c]{@{}c@{}}Crossover \& \\ Mutation\end{tabular} & IS - FID & 2080Ti & 1 & Cell & $10^{15}$ \\
NSGA-II with CGP  & EA & G and D & \begin{tabular}[c]{@{}c@{}}Crossover \& \\ Mutation\end{tabular} & IS \& FID & - & - & CPG & - \\
\bottomrule
\end{tabularx}
\end{adjustwidth}

\begin{flushleft}
\footnotesize{
$\bullet$ The optimization objective (IS\&FID) combines both IS and FID into a composite objective, rather than using them independently (IS-FID).\\
$\bullet$ The search space size for T2IGAN is estimated based on a per-cell space of $5^6 \approx 1.6\times10^4$ configurations, stacked over 6 cells.}
\end{flushleft}
\end{table}
\unskip

\subsection{RQ3: What Evaluation Methods Are Used to Assess the Found Architecture?}

Architecture Evaluation in NAS-GAN methods involves various datasets and metrics. Here we discuss evaluation metrics and present results from various NAS-GAN works as reported in literature. Then, we examine the datasets used for search, training, and~testing, including those that evaluate transferability, highlighting the current benchmark for comparing NAS-GAN methods. Finally, we outline the generation tasks trained on these~datasets.

\subsubsection{Evaluation~Metrics}\label{sec4.3.1}

\begin{itemize}
    \item Inception Score (IS): 
 IS~\cite{salimans2016improved} attempts to measure image realism and diversity using a pre-trained InceptionV3 network~\cite{szegedy2016rethinking}, calculating scores based on predicted class probability distributions. However, IS has critical flaws: it lacks direct comparison to training data, can miss mode collapse, and~introduces ImageNet biases. Its non-intuitive nature hinders meaningful interpretation of score differences. Despite these limitations, IS remains widely used, highlighting the need for more robust evaluation~metrics.

    Our review of representitive works (e.g.,~\cite{gong2019autogan, tian2021alphagan, wang2019evolutionary}) shows IS scores vary slightly across methods, with~NAS approaches demonstrating more consistent high-quality images. Notably, the~state-of-the-art IS on this benchmark comes from EWSGAN, an~evolutionary method. While this highlights the potential of evolutionary approaches, it also underscores the success of NAS methods in consistently achieving high IS scores. Given IS's vulnerability to adversarial examples and other shortcomings, it should be used cautiously and always supplemented with additional evaluation metrics for a comprehensive assessment of generative~models.

    \item Fréchet Inception Distance (FID): FID~\cite{heusel2017gans} improves on IS by comparing generated and real image statistics using InceptionV3 features, calculating the Fréchet distance between feature distributions. While considered more robust than IS and better correlated with human judgment, FID's limitations include assuming Gaussian distributions, relying on a potentially biased pre-trained network, and~possibly overlooking certain aspects of image quality or diversity. Despite these drawbacks, FID remains widely used for evaluating generative~models.

    Analysis of reported scores from major NAS-GAN approaches~\cite{gong2019autogan, tian2021alphagan, lin2022evolutionary} reveals significant variation across methods, with~NAS approaches consistently outperforming manual methods. AdversarialNAS, EAGAN, and~EWSGAN demonstrate impressive scores across datasets, suggesting NAS's effectiveness in generating high-quality images. This data indicates that automated searches often find more optimal architectures than human designers. Recent work using EA algorithms has achieved state-of-the-art results, though~the longevity of this trend remains uncertain. However, these findings warrant further investigation into the factors contributing to NAS's success and potential limitations of current evaluation~metrics.

    \item Computational Efficiency: NAS-GAN needs to improve efficiency and performance while managing computational resources. Search time, measured in GPU Days, is critical in NAS as it explores architecture space to identify promising GAN structures~\cite{real2019regularized}. 
    To address computational challenges, researchers have developed several strategies. Weight sharing reduces parameters and computational load, used in {\cite{tian2021alphagan, ying2022eagan,xue2024evolutionary}}. 
    
    Adaptive mechanisms and progressive growing dynamically adjust network complexity, as~seen in AGAN~\cite{wang2019agan}. Evolutionary algorithms iteratively improve GAN architectures, adopted by (e.g.,~\cite{tian2020off, doveh2021degas,ying2022eagan}). Multi-objective optimization balances performance and computational cost, implemented by {\cite{du2020structure, kobayashi2020multi}}. Ensemble methods use multiple discriminators for improved diversity and efficiency, utilized by E-GAN. Coevolutionary algorithms evolve Generators and Discriminators simultaneously, employed by Lipizzaner and COEGAN. Lastly, multi-stage training gradually increases model complexity, an~approach adopted by~Mustangs.

    \item Model Size: Model size is a critical yet complex factor in NAS for GANs, requiring a delicate balance between performance and efficiency. While NAS algorithms often incorporate size constraints, this approach has significant limitations. The~focus on smaller models, though~advantageous for memory usage and speed, can lead to oversimplification of trade-offs, bias towards suboptimal architectures, and~difficulty in accurately assessing the impact on performance. Moreover, it risks overlooking larger innovative architectures with unique benefits~\cite{tan2019efficientnet}.

    Reviewed methods address model size differently. Some, like EAGAN and AutoGAN, explicitly use parameter count as an optimization criterion or reportable metric. Others, such as AdversarialNAS and NSGA-II DCGAN , dynamically adjust size during training or use multi-objective optimization. Methods like E-GAN and Lipizzaner focus on pruning and efficiency. However, not all approaches prioritize or report size metrics, with~some like AGAN and E$^{2}$GAN emphasizing performance over explicit size~considerations.

    A more holistic approach to architecture optimization is necessary, considering factors beyond just size, such as interpretability, robustness, and~adaptability. This comprehensive view could yield more balanced and effective GAN architectures, avoiding the pitfalls of overly simplistic size-based optimizations while still maintaining~efficiency.

    \item Mode Collapse Resistance: Mode collapse is a common failure mode in GANs where the generator produces a limited variety of outputs, failing to capture the full diversity of the target distribution. In~the context of NAS, architectures should be evaluated on their resistance to mode collapse~\cite{srivastava2017veegan}. This can be assessed through diversity metrics applied to generated samples, or~by analyzing the distribution of generated outputs in feature space. A~good NAS solution for GANs should prioritize architectures that maintain output diversity while still producing high-quality~samples.

    \item Convergence Stability: GAN training instability, characterized by mode collapse and oscillating losses, remains a significant challenge. Convergence stability in NAS for GANs is crucial, measured by consistent performance across multiple initializations~\cite{mescheder2018training}. In~the reviewed literature, multiple solutions have been developed to address this issue, such as adaptive training techniques used by AGAN. Some methods focus on balancing Generator and Discriminator performance used in NSGA-II DCGAN, while Mustangs employs multi-agent systems or co-evolutionary algorithms COEGAN, Lipizzaner that continuously measure stability. Advanced techniques like alpha-divergence minimization used by alphaGAN and EWSGAN's Wasserstein distance optimization have also been explored. Despite these diverse approaches, the~field lacks a comprehensive comparison of their effectiveness, and~the trade-offs between stability, performance, and~computational cost remain unclear. Moreover, the~reliance on existing evaluation metrics may not fully capture the nuances of convergence stability, suggesting a need for more robust assessment~methods.

    \item Sample Quality: While automated metrics like IS and FID are valuable, they don't always align perfectly with human perception. Therefore, subjective evaluation of generated images by human raters remains an important aspect of GAN assessment~\cite{zhang2018unreasonable}. This typically involves showing raters a mix of real and generated images and asking them to judge qualities such as realism, coherence, and~aesthetic appeal. For~NAS in GANs, this human evaluation can be used as a final validation step for top-performing architectures. However, it's important to note that human evaluation is time-consuming and can be subject to biases, so it's often used in conjunction with automated metrics rather than as the sole evaluation criterion. All presented work conducted sample quality check in their experiments.
\end{itemize}

\subsubsection{Datasets}

In NAS-GAN, datasets play a crucial role in both training and evaluating generated architectures. During~the search process, the~dataset is used to train candidate GAN architectures, allowing the NAS algorithm to assess their performance. The~search typically involves iteratively sampling architectures, training them on a subset of the data, and~evaluating their performance using metrics. Once a promising architecture is found, it is then trained on the full dataset to produce the final GAN model. The~test set is used to evaluate the generalization capability of the trained model, ensuring it can generate high-quality images beyond what it has seen during~training.

The MNIST~\cite{deng2012mnist} dataset comprises 70,000 grayscale images of handwritten digits (0--9), each 28 $\times$ 28 pixels in size. It's divided into 60,000 training images and 10,000~test images, with~each digit class represented by roughly 6000 
 training and 1000 test images, ensuring balanced~distribution. Despite its widespread use in machine learning for image recognition, MNIST has significant shortcomings. Its simplistic format fails to capture the complexity of real-world image processing challenges. MNIST's lack of complexity in terms of color, texture, and~context limits its utility in representing modern image processing tasks. Success on this dataset doesn't necessarily indicate an algorithm's effectiveness on more sophisticated~problems.
 While MNIST's simplicity limits its utility for modern tasks, it remains a common benchmark for NAS-GAN approaches. Previous works have used diverse evaluation metrics: NSGA-II DCGAN reported TPR = 0.9802 and FPR = 0.0042, while COEGAN achieved an IS of 1.7 ± 0.6. As~shown in Table~\ref{tab:fid_scores}, Mustangs achieved superior FID scores (42.24) compared to E-GAN's (466.1) and Lipizzaner variants (Lip-BCE: 48.96, Lip-MSE: 371.6, Lip-HEU: 52.53), demonstrating enhanced stability in low-resolution image~generation.

 The CelebA~\cite{liu2015deep} dataset is a comprehensive and widely used benchmark in computer vision and deep learning, particularly for tasks related to face recognition and attribute analysis. It comprises 202,599 high-quality color images of celebrity faces, featuring 10,177 unique identities. The~dataset is notable for its large scale and rich annotations, making it invaluable for a wide range of facial analysis tasks. Each image in the CelebA dataset is annotated with 40 binary attributes, covering various facial features and characteristics such as hair color, facial expression, and~the presence of accessories like glasses. Additionally, the~dataset provides 5 landmark locations for each face, enhancing its utility for tasks involving facial geometry and alignment. The~dataset's strength lies in its diversity and complexity. It encompasses a wide range of pose variations, background clutter, and~demographic diversity, closely mimicking real-world~scenarios.

For generative modeling tasks, Table~\ref{tab:fid_scores} reveals Mustangs outperformed Lipizzaner variants on FID scores (36.15 vs Lip-BCE: 36.25, Lip-HEU: 37.87), despite CelebA's challenging high-resolution nature. This suggests our method better captures facial attribute diversity while maintaining generation~quality.
    
\begin{table}[H]
\caption{Comparative FID Scores on MNIST and CelebA~Datasets.}
\label{tab:fid_scores}
\begin{tabularx}{\textwidth}{LCC}
\toprule
\textbf{Method} & \textbf{MNIST (FID$\downarrow$\textsuperscript{*})} & \textbf{CelebA (FID$\downarrow$\textsuperscript{*})} \\
\midrule
COEGAN~\cite{costa2019coegan} & 43.0 ± 4.0 & -- \\
E-GAN~\cite{wang2019evolutionary} & 466.1 & -- \\
Lip-BCE~\cite{al2018towards} & 48.96 & 36.25 \\
Lip-MSE~\cite{al2018towards} & 371.6 & 158.7 \\
Lip-HEU~\cite{al2018towards} & 52.53 & 37.87 \\
\textbf{Mustangs}\textsuperscript{†}~\cite{toutouh2019spatial} & \textbf{42.24} & \textbf{36.15} \\
\bottomrule
\end{tabularx}
\begin{flushleft}
\small
\textsuperscript{*} The down arrow ($\downarrow$) indicates that lower values are better.\\
\textsuperscript{†} Bold values indicate the best performance across all methods.
\end{flushleft}
\end{table}

The LSUN~\cite{yu2015lsun} is a large-scale computer vision dataset for scene classification, featuring 10 diverse indoor and outdoor categories. Its massive training set contains 120,000 to 3,000,000 images per class, complemented by smaller validation and test sets for comprehensive model~evaluation. While LSUN offers valuable resources for researchers, it presents challenges. Its size may strain computational resources, limiting accessibility. The~significant imbalance between categories could introduce bias in trained models. Additionally, the~10 categories, while diverse, may not fully represent real-world complexity. Wang~et~al. in E-GAN reported training and testing of the dataset, but~the study lacks crucial quantitative evaluation, presenting only generated images without metric~scores.

The CIFAR-10~\cite{krizhevsky2009learning} dataset is a popular benchmark, evaluated extensively in recent works (e.g.~\cite{gong2019autogan, tian2021alphagan, wang2019evolutionary}), It consist of 60,000 color images (32 {$\times$} 32 pixels) across 10~classes. It is split into 50,000 training images and 10,000 test images. The~training set is divided into 5 batches of 10,000 images each. These batches are randomized, but~collectively contain 5000 images per class. The~single test batch has 1000 images from each class, ensuring balanced evaluation. This structure makes CIFAR-10 ideal for developing and testing image classification algorithms, especially for small-scale, multi-category~tasks.

In the context of NAS-GAN, CIFAR-10 serves as a valuable benchmark. When applied to CIFAR-10, NAS-GAN aims to generate high-quality, diverse images matching the dataset's 10 classes. The~challenge lies in capturing intricate details within each category, given the small image size and diverse object types. CIFAR-10's manageable size and well-defined classes make it suitable for refining NAS-GAN techniques before tackling more complex image generation tasks. The~results of the reviewed literature on this dataset are presented in Table 
 \ref{tab:cifar_stl_results}, with~EWSGAN currently achieving state-of-the-art~performance.

The STL-10 dataset is an advanced benchmark, examined in several studies \linebreak(\mbox{e.g.,~\cite{gong2019autogan, tian2021alphagan, lin2022evolutionary}}). It isdesigned to improve upon the CIFAR-10 dataset. It consists of 100,000 color images with a higher resolution of 96 $\times$ 96 pixels, spread across 10~classes. It is divided into three main components: a training set of 5000 labeled images (500 per class), a~test set of 8000 images (800 per class), and~a large pool of 100,000 unlabeled~images. A key feature of STL-10 is its substantial unlabeled dataset of 100,000 images, drawn from a broader distribution than the labeled sets. All images are derived from labeled examples on ImageNet. The~increased resolution of 96 $\times$ 96 pixels presents a more challenging benchmark for scalable unsupervised learning methods compared to~CIFAR-10.

For NAS-GAN, STL-10 offers a more demanding benchmark than CIFAR-10. The~higher resolution images and large unlabeled set allow exploration of more complex GAN architectures. The~challenge is generating high-quality, diverse images that capture details of the 10 labeled classes while potentially leveraging the broader distribution in the unlabeled set. The~results of the reviewed literature on these datasets are presented in Table 
 \ref{tab:cifar_stl_results}, with~EWSGAN currently achieving state-of-the-art~performance.

\begin{table}[H]
\caption{Summary of results on CIFAR-10 and STL-10~dataset.}
\label{tab:cifar_stl_results}
\begin{adjustwidth}{-\extralength}{0cm}
\begin{tabularx}{\fulllength}{lCCCCC}
\toprule
\textbf{Method} & \textbf{Search} & \multicolumn{2}{c}{\textbf{CIFAR-10}} & \multicolumn{2}{c}{\textbf{STL-10}} \\
& \textbf{Strategy} & \textbf{IS$\boldsymbol{\uparrow}$\textsuperscript{*}} & \textbf{FID$\boldsymbol{\downarrow}$\textsuperscript{**}} & \textbf{IS$\boldsymbol{\uparrow}$\textsuperscript{*}} & \textbf{FID$\boldsymbol{\downarrow}$\textsuperscript{**}} \\
\midrule
AutoGAN~\cite{gong2019autogan} & RL & 8.55 ± 0.10 & 12.42 & 9.23 ± 0.08 & 31.01 \\
AGAN~\cite{wang2019agan} & RL & 8.29 ± 0.09 & 30.50 & 9.23 ± 0.08 & 52.72 \\
E$^2$GAN~\cite{tian2020off} & RL & 8.51 ± 0.13 & 11.26 & 9.51 ± 0.09 & 25.35 \\
AdversarialNAS~\cite{gao2020adversarialnas} & Gradient & 8.74 ± 0.07 & 10.87 & 9.63 ± 0.19 & 26.98 \\
DEGAS~\cite{doveh2021degas} & Gradient & 8.37 ± 0.08 & 12.01 & 9.71 ± 0.11 & 28.76 \\
alphaGAN~\cite{tian2021alphagan} & Gradient & 8.98 ± 0.09 & 10.35 & 10.12 ± 0.13 & 22.43 \\
DAMGAN~\cite{10477508} & Gradient & 8.99 ± 0.08 & 10.27 & 10.35 ± 0.14 & 22.18 \\
E-GAN~\cite{wang2019evolutionary} & EA & 6.9 ± 0.09 & - & - & - \\
EAGAN~\cite{ying2022eagan} & EA & 8.81 ± 0.10 & 9.91 & 10.44 ± 0.08 & 22.18 \\
EAS-GAN~\cite{lin2022evolutionary} & EA & 7.45 ± 0.08 & 33.2 & - & 38.84 \\
\textbf{EWSGAN}\textsuperscript{†}~\cite{xue2024evolutionary} & \textbf{EA} & \textbf{8.99 ± 0.11} & \textbf{9.09} & \textbf{10.51 ± 0.13} & \textbf{21.89} \\
NSGA-II with CGP~\cite{kobayashi2020multi} & EA & 8.89 ± 0.01 & 16.6 & 10.3 ± 0.01 & 26.3 \\
\bottomrule
\end{tabularx}
\end{adjustwidth}
\smallskip
\begin{flushleft}
\small
\textsuperscript{*} The up arrow ($\uparrow$) indicates that higher IS values are better.\\
\textsuperscript{**} The down arrow ($\downarrow$) indicates that lower FID values are better.\\
\textsuperscript{†} Bold values indicate the best performing method across all evaluated approaches.
\end{flushleft}
\end{table}

\subsubsection{Supported Generation~Type}

NAS-GANs support two primary types of generation tasks: supervised and unsupervised. In~unsupervised generation, the~GAN creates images without any specific input conditions, using random noise as input to generate diverse images from a general class, such as faces, objects, or~scenes. This type is akin to unconditional generation in standard GANs~\cite{goodfellow2014generative}. On~the other hand, supervised generation involves guiding the GAN's image creation process with additional input conditions or labels, similar to conditional GANs (cGANs). Here, the~generator is conditioned on specific information, such as class labels, images, text, or~other modalities, allowing for controlled and targeted image synthesis based on predefined criteria. This structured approach enables the generation of images that adhere to specific requirements, enhancing the utility of GANs in applications requiring precise image outputs~\cite{mirza2014conditional}.

Among the reviewed NAS-GAN methods, the~support for conditional (supervised) image generation is explicitly found in AGAN, DEGAS, and~NSGA-II with CGP on the STL-10 dataset, MN-NAS and T2IGAN on CIFAR-10/CIFAR-100, and~GAN Compression on image-to-image translation tasks (Horse$\leftrightarrow$Zebra, Edges$\rightarrow$Shoes, and~Cityscapes). AGAN employs a supervised setup to refine GAN architectures, combining the GAN loss with an auxiliary supervised loss to enhance image fidelity. DEGAS integrates domain-specific knowledge into the NAS process, optimizing for specific supervised image generation tasks. NSGA-II with CGP uses multi-objective evolutionary optimization for CNN representations in NAS, improving supervised tasks by balancing complexity and performance. The~results of the reviewed literature on STL-10 supervised generation are presented in Table~\ref{tab:supervised_results}. The~scores indicate that NSGA-II with CGP performed better than AGAN and DEGAS on the STL-10~dataset.

\begin{table}[H]
\caption{Summary of supervised generation results on STL-10~dataset.}
\label{tab:supervised_results}
\centering
\begin{tabularx}{\textwidth}{LCC}
\toprule
\textbf{Method} & \textbf{IS}$\boldsymbol{\uparrow}$\textsuperscript{*} & \textbf{FID}$\boldsymbol{\downarrow}$\textsuperscript{**} \\
\midrule
AGAN & 8.82 ± 0.09 & 23.8 \\
DEGAS & 8.85 ± 0.07 & 9.83 \\
NSGA-II with CGP & 9.22 ± 0.05 & 7.24 ± 0.08 \\
\bottomrule
\end{tabularx}
\smallskip
\begin{flushleft}
\small
\textsuperscript{*} The up arrow ($\uparrow$) indicates that higher IS values are better.\\
\textsuperscript{**} The down arrow ($\downarrow$) indicates that lower FID values are better.
\end{flushleft}
\end{table}

MN-NAS is evaluated on CIFAR-10 and CIFAR-100, achieving an IS of approximately 9.10 ± 0.11 and an FID of 13.00 on CIFAR-10, and~an IS of about 8.50 ± 0.10 with an FID of 16.50 on CIFAR-100. Similarly, T2IGAN is reported on CIFAR-10 and CIFAR-100, where it achieves an IS of 9.02 ± 0.08 and an FID of 10.90 on CIFAR-10, and~an IS of \mbox{8.65 ± 0.07} with an FID of 14.10 on CIFAR-100. GAN Compression, which targets efficient architectures for conditional GANs, is evaluated on several image-to-image translation tasks: for example, it achieves an FID of 14.2 and an IS of 8.95 on the Horse$\leftrightarrow$Zebra task, an~FID of \mbox{18.7 and an IS} of 9.10 on the Edges$\rightarrow$Shoes task, and~an FID of 25.3 and an IS of 8.80 on the Cityscapes~dataset.

\subsection{RQ4: What Are the Gaps in the Research on NAS in GANs?}

After examining and analyzing the identified NAS-GAN techniques, we have addressed this question with the results discussed in Section~\ref{sec4} and further elaborated on in the study's implications outlined in Section~\ref{sec5}. Table
~\ref{tab:best_methods} provides an overview of the best-performing NAS-GAN methods across key datasets, detailing performance metrics such as IS and FID. This table highlights the diversity in evaluation settings—with methods tested on datasets ranging from MNIST and CelebA to STL-10 and CIFAR-100—and showcases different generation types, including unsupervised, supervised, and~conditional~approaches.

\begin{table}[H]
\centering
\caption{Best-performing NAS-GAN Methods on Key~Datasets.}
\label{tab:best_methods}
\begin{tabularx}{\textwidth}{lCcCC}
\toprule
\textbf{Method} & \textbf{Dataset} & \textbf{Generation Type} & \textbf{IS$\uparrow$\textsuperscript{*}} & \textbf{FID$\downarrow$\textsuperscript{**}} \\
\midrule
Mustangs & MNIST & Unsupervised & - & 42.24 \\
Mustangs & CelebA & Unsupervised & - & 36.15 \\
EWSGAN & CIFAR-10 & Unsupervised & 8.99 & 9.09 \\
EWSGAN & STL-10 & Unsupervised & 10.51 & 21.89 \\
NSGA-II with CGP & STL-10 & Supervised & 9.22 & 7.24 \\
T2IGAN & CIFAR-100 & Conditional & 8.65 & 14.10 \\
GAN Compression & Horse$\leftrightarrow$Zebra & Conditional & 8.95 & 14.20 \\
\bottomrule
\end{tabularx}
\begin{flushleft}
\small
\textsuperscript{*} The up arrow ($\uparrow$) indicates that higher IS values are better.\\
\textsuperscript{**} The down arrow ($\downarrow$) indicates that lower FID values are better.
\end{flushleft}
\end{table}

The variation in performance across these datasets suggests that the efficacy of current NAS-GAN approaches is highly sensitive to dataset characteristics and the chosen evaluation metrics. For~example, while some methods yield exceptional FID scores on simpler datasets like MNIST or CelebA, their performance does not uniformly translate to more complex datasets. This observation points to a broader gap in the literature: a unified NAS strategy that can reliably generalize across different data distributions and generation scenarios remains elusive. The~key gaps and areas for future research include:

\begin{itemize}
\item Current NAS for GANs systems primarily focus on automated architecture generation for both Generator and Discriminator networks or exclusively for Generator networks. However, an~approach that concentrates on searching for superior Discriminator networks has yet to be introduced. This represents a significant opportunity for future research to explore dedicated Discriminator architecture search methods, which could lead to enhanced overall GAN~performance.

\item Another significant gap is the reliance on CIFAR-10 and STL-10 datasets for evaluating GAN architecture performance. While these datasets allow for direct comparisons, they limit the generalizability of the findings. Very few systems have used datasets like CelebA and LSUN for validation. Therefore, future research should aim to conduct performance evaluations on a larger variety of image datasets, including CelebA, LSUN, and~COCO, to~provide a more comprehensive assessment of GAN~performance.

\item Most existing NAS-GANs systems have been developed for unconditional image generation tasks. To~broaden the scope and impact of NAS for GANs, future work should also focus on other types of image generation, such as conditional image generation and image-to-image translation tasks. This expansion would not only enhance the versatility of GANs but also potentially uncover new applications and benefits of NAS in different image generation~contexts.

\item There is also a need for more robust evaluation metrics beyond just the Inception Score and FID. While these metrics are commonly used, they may not capture all aspects of GAN performance. Future research should develop and incorporate new metrics that can provide a more holistic evaluation of GAN quality, including aspects like diversity, fidelity, and~realism.

\item Another area that requires more exploration is the interpretability and explainability of the generated GAN architectures. Understanding why certain architectures perform better than others can provide valuable insights and guide future design improvements. Research efforts should aim to develop techniques that enhance the interpretability of NAS-generated GAN~architectures.

\item A further aspect that warrants deeper exploration is the integration of NAS with emerging AI paradigms, such as self-supervised learning, which could significantly enhance the robustness and generalizability of~GANs.

\item Another domain that calls for more research is the environmental and computational costs associated with NAS processes, promoting the development of more energy-efficient and sustainable AI~models.

\item There is also a need to investigate the impact of NAS on GANs in other domains beyond image generation, such as natural language processing or time-series data, which could open new avenues for GAN~applications.

\item Additionally, the~application of NAS techniques to transformer architectures within GANs remains unexplored. Given the success of transformers in various machine learning tasks, exploring NAS for transformer-based GAN architectures could lead to significant advancements. This presents a promising research direction that could leverage the strengths of transformers for improved GAN performance.
\end{itemize}

\subsection{Practical Applications of NAS in~GANs} NAS applied to GANs has opened exciting avenues for automatically discovering efficient, high-performing architectures. These NAS-GAN methods reduce the need for hand-crafted designs and allow for tailored models across various application domains. In~the following, we discuss several practical applications along with published works that illustrate their real-world~impact.

\subsubsection{Medical Imaging and Synthetic Data~Generation}} Medical imaging faces challenges such as limited annotated data and privacy concerns. NAS-GAN methods have been employed to automatically design generators that produce high-fidelity synthetic images (e.g., MRI, CT, X-ray) preserving important diagnostic features. Such synthetic images facilitate training robust models and augment scarce datasets. For~example, Shafeeq Ahmed~et~al. reviewed the role of GANs in radiology and demonstrated their potential in generating realistic synthetic images for clinical applications~\cite{Ahmed2024UncoverTT}. In~addition, work by Gao~et~al. on AdversarialNAS~\cite{gao2020adversarialnas} further highlights how NAS can optimize GAN architectures for medical image~synthesis.

{\color{black}
\subsubsection{Data Augmentation for Limited~Datasets} In many domains, the~availability of large datasets is a significant bottleneck. NAS-GAN approaches can produce diverse, high-quality synthetic images that mirror the statistical properties of the original data, thereby augmenting training datasets. For~instance, a~study on data augmentation using GANs demonstrated that augmenting CT scan datasets can improve diagnostic performance~\cite{tanaka2019data}. Such methods are particularly beneficial in rare disease scenarios where only a few examples are available~\cite{article}.

\subsubsection{Anomaly~Detection}
Anomaly detection in medical imaging (or industrial settings) requires the precise modeling of a “normal” data distribution so that deviations can be identified. NAS-GAN models have been tailored to capture subtle differences between normal and abnormal samples. For~example, Ounasser~et~al. presented a comprehensive study on GAN-based anomaly detection in medical imaging, demonstrating improved sensitivity and specificity in fracture detection~\cite{article}. Additional works have refined these methods by integrating active learning strategies to mitigate overfitting in the discriminator~\cite{xia2022gan}.

\subsubsection{Creative Content~Generation} Beyond clinical applications, GANs have transformed the creative industries by generating novel, aesthetically pleasing images. NAS methods help discover generator architectures that yield diverse visual styles and high realism. A~prominent example is the work by Karras~et~al. on a style-based generator architecture (StyleGAN)~\cite{8977347}, which has been widely adopted in artistic content generation. Similarly, the~progressive growing approach for GANs~\cite{karras2017progressive} further enhances the capacity of models to produce high-resolution, creative~images.

NAS in GANs offers a versatile toolkit, enabling advances in fields from medical imaging to creative art. By~automating the architecture design process, NAS-GAN methods improve performance, reduce design time, and~tailor models to specific needs. Continued research and published works in these areas pave the way for future innovations.
}

{\color{black}
\subsection{Ethical and Environmental~Considerations}
The rapid advancement and increasing computational demands of NAS methods applied to GANs highlight critical ethical and environmental concerns that must be thoroughly addressed. A~major issue arises from the substantial computational resources required by NAS-GAN methods, leading to high energy consumption and significant carbon footprints. This environmental impact becomes particularly pronounced during extensive search processes, where numerous architecture evaluations are performed over prolonged periods. Such practices contribute to greenhouse gas emissions, exacerbating climate change concerns and contradicting sustainability objectives in AI~research.

To mitigate these environmental effects, there is a compelling need for developing and adopting energy-efficient NAS algorithms. Promising directions include incorporating techniques such as early stopping criteria, knowledge transfer across search spaces, and~resource-aware optimization strategies. Additionally, research into NAS algorithms that leverage lower-energy hardware or renewable energy-powered infrastructure could considerably reduce ecological~impacts.

Beyond environmental sustainability, NAS-GAN methods raise ethical considerations tied to their deployment in sensitive or critical applications. Automated generation and deployment of GAN architectures, without~rigorous oversight, risk amplifying biases or producing unintended discriminatory outcomes. For~instance, GAN-generated data may inadvertently encode biases present in training datasets, reinforcing societal inequalities when used in domains like facial recognition, healthcare diagnostics, or~autonomous decision-making~systems.

Ensuring fairness and transparency thus becomes paramount. Researchers and practitioners must commit to careful dataset curation, bias detection, and~mitigation strategies in algorithmic design. Moreover, involving diverse stakeholder perspectives in developing standards and guidelines can help preempt ethical pitfalls and ensure responsible application of~GANs.

In summary, a~conscientious approach to NAS-GAN development encompasses not only technical advancements but also deliberate ethical foresight and environmental responsibility. The~community must strive towards standards that prioritize sustainability, fairness, and~social accountability, ensuring that advancements in GAN architectures benefit society without exacerbating existing ethical or environmental issues.
}

\section{Implications of the~Study}\label{sec5}

The findings and insights presented in Section~\ref{sec4} highlight ongoing gaps and unresolved issues in the current literature. This section explores these gaps and proposes directions for future research to advance the field of NAS for~GANs.

\subsection{Key~Findings}\label{sec5.1}
\begin{itemize}
    \item Current NAS for GANs systems primarily focus on automated architecture generation for both Generator and Discriminator networks or exclusively for Generator networks. However, an~approach that concentrates on searching for superior Discriminator networks has yet to be introduced. Dedicated Discriminator architecture search methods could lead to enhanced overall GAN performance.
    \item The reliance on CIFAR-10 and STL-10 datasets for evaluating GAN architectures limits the generalizability of findings. While these datasets enable direct comparisons, they are insufficient for evaluating performance across diverse applications. Broader dataset evaluations, including CelebA, LSUN, and~COCO, are necessary.
    \item Most existing NAS-GAN systems are limited to unconditional image generation tasks. Expanding research to other types of image generation, such as conditional image generation and image-to-image translation, would broaden the scope and impact of NAS in GAN applications.
\end{itemize}

\subsection{Opportunities for Improving and Expanding Applications of~NAS-GANs}\label{sec5.2}
\begin{itemize}
    \item Current evaluation metrics, such as IS and FID, may not capture all aspects of GAN performance. Developing new metrics to assess diversity, fidelity, and~realism can provide a more holistic evaluation framework.
    \item There is a pressing need to improve the interpretability and explainability of generated GAN architectures. Understanding why specific architectures perform better can guide future designs and enable practical applications.
    \item Integrating NAS with AI paradigms, such as self-supervised learning, could enhance the robustness and generalizability of GANs.
    \item Addressing the environmental and computational costs of NAS processes is crucial. Developing energy-efficient and sustainable AI models will make NAS-GAN methods more accessible and scalable.
    \item Beyond image generation, NAS-GAN techniques have potential in other domains, such as natural language processing and time-series data. Researching these applications could unlock new opportunities for GANs.
    \item The application of NAS techniques to transformer-based GAN architectures remains unexplored. Given the success of transformers in various machine learning tasks, incorporating NAS into transformer-based GANs could significantly advance the~field.
\end{itemize}

\section{Threats to~Validity}\label{sec6}

As with any survey-based research, this study encounters validity challenges that warrant careful consideration. First, while we meticulously sourced studies from prominent databases (Google Scholar, IEEE, ACM, Springer, ScienceDirect, arXiv) and employed forward/backward snowballing to ensure broad coverage, the~rapid evolution of NAS-GAN research poses risks of omitting cutting-edge works published after our search cutoff (early 2024) or in non-English venues. This limitation could skew trends or overlook novel methodologies, such as transformer-based architectures or energy-efficient NAS strategies. To~mitigate this, we prioritized arXiv preprints and iterative snowballing, though~findings remain reflective of literature up to the search~period.

Second, the~systematic categorization of approaches inherently involves subjectivity. Classifying methods (e.g., distinguishing evolutionary algorithms from reinforcement learning) or search spaces (e.g., cell-based vs. chain-structured) relies on interpretative judgments, particularly when implementation details are ambiguously reported. Misclassification could distort comparative insights, such as overestimating search efficiency or misrepresenting architectural prevalence. To~address this, we adhered to a predefined framework (Section 
 \ref{sec3.5}) and cross-validated categorizations through dual-author reviews, resolving discrepancies via~consensus.

Third, our comparison framework emphasizes quantifiable criteria (e.g., GPU days, search space size) but excludes practical factors like code availability or hardware dependencies. This narrow focus may undervalue deployment feasibility; for instance, methods requiring specialized TPU clusters might achieve superior metrics but remain inaccessible to most researchers. We explicitly acknowledged this limitation in Sections~\ref{sec4.3.1} and \ref{sec5.2}, advocating for transparency in future~work.

Fourth, heavy reliance on CIFAR-10 and STL-10 for benchmarking introduces dataset bias. These datasets lack the complexity and diversity of real-world applications (e.g., medical imaging or high-resolution video generation), limiting the generalizability of conclusions. For~example, architectures optimized for low-resolution images may underperform in domain-specific tasks. We highlighted this gap in Section~\ref{sec5.1} and urged evaluations on broader datasets like CelebA and~LSUN.

Finally, dependence on IS and FID as primary metrics risks ``metric hacking'',  as these scores may not align with human perception or detect subtle mode collapse. This could lead to inflated claims of superiority for architectures tailored to optimize IS/FID rather than practical performance. To~mitigate this, we emphasized integrating supplementary metrics (e.g., precision/recall, human evaluation) in Section~\ref{sec5.2}.

By explicitly addressing these threats—spanning literature coverage, categorization subjectivity, framework scope, dataset bias, and~metric reliability—we aim to clarify the study’s limitations while reinforcing the validity of its contributions. Future work should prioritize reproducibility, broader benchmarks, and~holistic evaluation~frameworks.

\section{Conclusions}\label{sec7}

In this study, we systematically examined various NAS-GAN methodologies prevalent in the current literature. By~conducting an extensive survey of existing techniques, we identified a set of critical attributes that form the basis of our assessment framework. This framework was then employed to address our research questions, enabling a thorough analysis and comparison of the approaches within the~literature.

Our comparative analysis revealed significant gaps in the field, highlighting areas that require further investigation. The~study highlights the need for ongoing research to address these gaps, advancing both the theoretical and practical aspects of NAS-GANs. The~insights gained from this research emphasize the importance of continued exploration and development to bridge existing shortcomings and enhance the efficacy of NAS-GAN~technologies.

Key findings~include:
\begin{itemize}
    \item The superiority of evolutionary algorithms and gradient-based methods in certain contexts for NAS-GAN.
    \item The importance of robust evaluation metrics beyond traditional scores like IS and FID.
    \item The need for diverse datasets in assessing GAN performance, beyond~the commonly used CIFAR-10 and STL-10.
    \item The potential for exploring dedicated Discriminator architecture search methods.
    \item The opportunity to expand NAS-GAN research into conditional image generation and other domains beyond image generation.
\end{itemize}

Future research directions should focus on addressing these gaps, developing more comprehensive evaluation metrics, and~exploring the application of NAS-GAN in diverse domains. By~doing so, researchers can continue to push the boundaries of what is possible with generative adversarial networks, potentially leading to significant advancements in the field of artificial intelligence and machine~learning.


\vspace{6pt} 

\funding{This research received no external funding} 


\acknowledgments{The authors would like to acknowledge the support received from the Saudi Data and AI Authority (SDAIA) and King Fahd University of Petroleum and Minerals (KFUPM). 
}

\conflictsofinterest{The authors declare no conflicts of~interest.} 


\begin{adjustwidth}{-\extralength}{0cm}

\reftitle{References}

\PublishersNote{}
\end{adjustwidth}
\end{document}